\documentclass[conference]{IEEEtran}

\usepackage{lettrine}
\usepackage{xcolor}
\usepackage{graphicx}
\usepackage{booktabs}
\usepackage{amsmath}
\usepackage{array}
\usepackage{tabularx}
\usepackage{hyperref}

\newcolumntype{L}[1]{>{\raggedright\arraybackslash}m{#1}}

\title{Facial Affect Analysis for Service-Oriented Systems: Advances, Challenges, and Future Visions}

\author{
    \IEEEauthorblockN{
        Spyridon Georgiou\IEEEauthorrefmark{1},
        Aggelos Psiris\IEEEauthorrefmark{1},
        Thomas Lagkas\IEEEauthorrefmark{2},
        Vasileios Argyriou\IEEEauthorrefmark{3},
    }
    \IEEEauthorblockN{
        Panagiotis Sarigiannidis\IEEEauthorrefmark{4},
        Iraklis Varlamis\IEEEauthorrefmark{1},
        Georgios Th. Papadopoulos\IEEEauthorrefmark{1}
    }
    \IEEEauthorblockA{
        \IEEEauthorrefmark{1}Department of Informatics and Telematics, Harokopio University of Athens, Athens, Greece\\
        \IEEEauthorrefmark{2}Department of Informatics, Faculty of Science, Democritus University of Thrace, Kavala, Greece\\
        \IEEEauthorrefmark{3}Department of Networks and Digital Media, Kingston University, Kingston upon Thames, United Kingdom\\
        \IEEEauthorrefmark{4}Department of Electrical and Computer Engineering, University of Western Macedonia, Kozani, Greece
    }
    \IEEEauthorblockA{
        Emails: \{sgeorgiou,aggelospsiris,varlamis,g.th.papadopoulos\}@hua.gr, tlagkas@cs.duth.gr,\\vasileios.argyriou@kingston.ac.uk, psarigiannidis@uowm.gr
    }
}

\begin{document}
\maketitle

\begin{abstract}
Facial Affect Analysis (FAA) is evolving from a stand-alone recognition task into a reusable perception capability for Service-Oriented Software Ecosystems (SoSE). This paper preserves the FAA methodological core while reframing recent advances through systems-engineering requirements for composable and dependable services. We review representative progress in static and dynamic expression analysis, action-unit and micro-expression modeling, and modern CNN, Transformer, graph, and hybrid architectures, then interpret these advances by their operational fit in edge, cloud, and hybrid service pipelines. The synthesis emphasizes SoSE concerns that determine deployability: service contracts for uncertainty-aware outputs, latency and availability envelopes, lifecycle monitoring and recalibration, governance-aware integration, and interoperability across independently evolving components. Our analysis shows that benchmark gains alone are insufficient for SoSE readiness; robustness under shift, intervention stability, fairness, privacy posture, and runtime guarantees are equally critical. We conclude with a roadmap for treating FAA as an operational service component with explicit interfaces, measurable quality attributes, and accountable lifecycle management.
\end{abstract}

\begin{IEEEkeywords}
Facial Affect Analysis, Facial Expression Recognition, Service-Oriented Software Ecosystems, Edge-Cloud AI, Uncertainty-Aware Services, Human-Centered AI, Runtime Monitoring, Trustworthy AI
\end{IEEEkeywords}

\section{Introduction}
\label{sec:introduction}
\lettrine{F}{acial} Affect Analysis has evolved from a stand-alone recognition problem into an operational component of modern digital services. In service-oriented systems, FAA outputs are not consumed as final results; they become inputs to orchestration logic, recommendation engines, risk monitoring pipelines, and adaptive user interfaces. This transition is already visible in smart learning platforms, mobile wellbeing systems, smart-home workflows, and driver-state services \cite{savchenko2022classifying,guo2022facial,islam2024facepsy,kaushik2022isecurehome,gu2024emotake}.

At the same time, the research community has produced substantial progress in deep FER, dynamic expression analysis, micro-expression recognition, and graph-based modeling \cite{li2020deep,li2022deep,liu2022graph}. These advances are important, but they are often discussed in a benchmark-centered language that does not fully capture service constraints. For deployed services, small gains on a closed dataset can be less important than stability under camera variation, acceptable latency under edge budgets, or predictable behavior across user populations.

This paper addresses that mismatch. Rather than replacing the FAA core, we reinterpret recent FAA literature through a service-oriented lens. The objective is to explain how methodological choices influence real-world service behavior, and why some technically strong models remain difficult to deploy.

From a Service-Oriented Systems Engineering perspective, FAA should be treated as a reusable service capability with explicit contracts, lifecycle monitoring, and governance constraints. In this framing, the central question is not only which model is most accurate, but whether FAA services remain composable, observable, and dependable inside larger orchestration pipelines. This includes defining how uncertainty is exposed to downstream services, how latency and availability targets are enforced under edge-cloud operation, and how fairness and privacy controls are validated during runtime. Positioning FAA in this way links computer-vision progress to system-level engineering outcomes such as quality of service, intervention stability, and accountable adaptation.

The main contributions are as follows:
\begin{itemize}
    \item we synthesize representative recent FAA advances, emphasizing methods with clear service relevance;
    \item we organize FAA capabilities into task, architecture, and deployment layers that align with service engineering decisions;
    \item we review how FAA is integrated in education, healthcare, smart environments, transport, and assistive systems;
    \item we provide a service-aware evaluation perspective and discuss deployment-critical future directions.
\end{itemize}

The remainder of the paper is organized as follows. Section~\ref{sec:surveymethodology} defines the review scope and positioning. Section~\ref{sec:methods} presents FAA capability layers for service-oriented systems, and Section~\ref{sec:task_advances} discusses task-specific advances through a deployment lens. Section~\ref{sec:applications} reviews representative service-oriented application domains. Section~\ref{sec:deployment_patterns} examines deployment architectures, integration challenges, and SoSE-oriented evaluation protocols, including lifecycle operations and service-aware metrics. Section~\ref{sec:implementation_guidelines} provides practical implementation guidelines, domain playbooks, and a worked deployment example. Section~\ref{sec:roadmap} summarizes future visions and open problems, and Section~\ref{sec:conclusions} concludes the paper.

\section{Review Scope and Positioning}
\label{sec:surveymethodology}
This review follows the PRISMA reporting principles \cite{page2021prisma} and provides a conceptual synthesis of recent FAA work relevant to service design and deployment, rather than a full systematic meta-analysis. Search terms combined FAA and service-oriented concepts, including \emph{facial expression recognition}, \emph{micro-expression recognition}, \emph{action unit detection}, \emph{smart education}, \emph{mobile sensing}, \emph{driver monitoring}, \emph{assistive services}, \emph{edge inference}, and \emph{cloud applications}, across major digital libraries (Scopus, IEEE Xplore, ACM Digital Library) and recent high-impact venues.

Included studies fall into three broad classes: modern surveys of FAA tasks and architectures \cite{li2020deep,li2022deep,liu2022graph,saadi2024driver}, representative methodological advances in robustness, temporal modeling, transfer, and efficiency \cite{li2022towards,fan2020facial,xue2022vision,zheng2023poster,liu2023expression,xu2024multiscale,chen2024static}, and application studies that embed FAA in practical service pipelines \cite{savchenko2022classifying,guo2022facial,islam2024facepsy,lyu2024emooly,lyu2024dailyconnect,kaushik2022isecurehome,gu2024emotake}. Excluded papers mainly covered older foundational affect theory, leaderboard-only comparisons without deployment insight, or task framings with limited transfer to service adaptation.

A central step in this survey is to preserve FAA methodological depth while clarifying what is missing in prior survey coverage. Recent influential reviews already offer strong technical categorization. However, most of them are organized around task taxonomies, architecture families, or domain-specific performance summaries. They are less explicit about service integration pathways, deployment constraints, and operational reliability requirements that become dominant outside controlled benchmarks.

For example, the deep FER review line synthesized representation learning advances and benchmark trends, but focused primarily on recognition modeling \cite{li2020deep}. Micro-expression surveys similarly provided detailed algorithmic coverage while emphasizing data scarcity and subtle motion modeling rather than deployment behavior \cite{li2022deep}. Graph-centered reviews illuminated structural modeling decisions but did not center service orchestration concerns \cite{liu2022graph}. Application-focused surveys, such as driver monitoring overviews, moved closer to deployment but remained domain-specific \cite{saadi2024driver}.

The present paper positions itself as a bridge: it does not replace those technical surveys, but reinterprets their key evidence through service-oriented integration questions. This includes deciding where lightweight static models are sufficient, where temporal pipelines are necessary, and where privacy or fairness constraints dominate architecture choices.

To make this positioning explicit, the review organizes evidence around service life-cycle concerns rather than benchmark ranking alone:
\begin{itemize}
    \item \textbf{Sensing and perception readiness:} what FAA tasks are stable enough for continuous service operation;
    \item \textbf{Orchestration compatibility:} how FAA outputs can be consumed by policy engines, recommendation modules, and monitoring workflows;
    \item \textbf{Deployment mode fit:} what changes when inference is cloud-based, edge-first, or hybrid;
    \item \textbf{Operational risk controls:} how uncertainty, fairness, privacy, and latency constraints shape safe adaptation;
    \item \textbf{Service impact validity:} whether reported gains translate into more stable and useful interventions in real contexts.
\end{itemize}

This service-oriented framing uses the same recent methodological evidence reviewed in prior FAA surveys, but evaluates it through deployability and operational trust. As a result, the paper emphasizes not only what models predict, but also when those predictions are reliable enough to drive service actions.

\section{FAA Capabilities for Service-Oriented Systems}
\label{sec:methods}
For service-oriented systems, FAA is best interpreted as a layered capability stack: task-level inference, architecture-level modeling, and deployment-level operation.

\begin{figure*}[t]
\centering
\includegraphics[width=\textwidth]{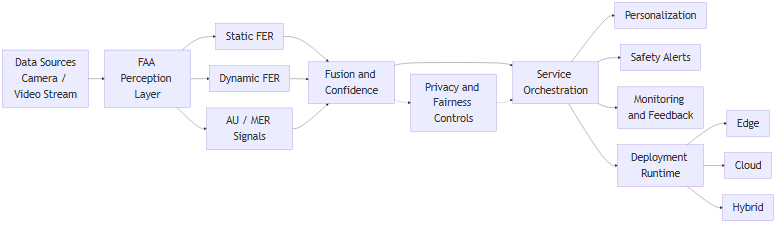}
\caption{Compact architecture for FAA-enabled service-oriented systems.}
\label{fig:section3_architecture}
\end{figure*}

Figure~\ref{fig:section3_architecture} summarizes this stack as an end-to-end service pipeline, from data sources and FAA perception modules to confidence-aware fusion, orchestration, and deployment runtime choices.

\subsection{Task Layer}
In Fig.~\ref{fig:section3_architecture}, the task layer corresponds to the \emph{FAA Perception Layer} and its three outputs (static FER, dynamic FER, and AU/MER signals) that are passed to \emph{Fusion and Confidence}. Static facial expression recognition remains the most deployable FAA capability for low-latency services. Recent methods improved performance by combining stronger supervision strategies with attention-based feature learning, reducing sensitivity to limited labels and noisy class boundaries \cite{li2022towards,fan2020facial}. In service terms, static FER is valuable because it can run frequently with modest compute and provide immediate adaptation signals.

Dynamic facial expression recognition adds temporal evidence and is more suitable for services that need continuity, such as engagement tracking and driver-state support. Recent sequence-aware designs improved short-term and long-term temporal reasoning \cite{liu2023expression,xu2024multiscale}. However, these gains come with higher computational and synchronization demands.

Micro-expression and action-unit centered pipelines provide finer-grained affect cues and can improve interpretability in sensitive applications. Still, they remain data-constrained and harder to stabilize in unconstrained deployment settings \cite{li2022deep,kollias2023multi}. For service architecture, they are often best used as specialized modules rather than universal defaults.

\subsection{Architecture Layer}
The architecture layer is reflected in how the perception branches and \emph{Fusion and Confidence} block are implemented in Fig.~\ref{fig:section3_architecture}. CNN-based systems continue to be attractive for edge and mobile services because they capture local facial structure efficiently. Transformer-based methods improve global dependency modeling and can better handle complex appearance variation, but often require larger compute budgets and data support \cite{xue2022vision,zheng2023poster}. Graph-based and hybrid systems are useful when explicit relationships among facial regions or action units matter \cite{liu2022graph}. In service contexts, architecture choice should therefore be treated as a constrained optimization problem rather than a single-model race.

\subsection{Deployment Layer}
Deployment mode shapes FAA utility, represented in Fig.~\ref{fig:section3_architecture} by the \emph{Deployment Runtime} branch into edge, cloud, and hybrid execution, and by the feedback coupling with privacy and fairness controls. Cloud-backed services can support heavier models and centralized analytics, but raise communication and privacy concerns \cite{guo2022facial}. On-device and edge pipelines improve responsiveness and data minimization, which is critical for personal and assistive services \cite{islam2024facepsy,lyu2024dailyconnect}. Cyber-physical services in homes or vehicles require stable low-latency behavior under environmental shift \cite{kaushik2022isecurehome,gu2024emotake}. These constraints should directly influence both model selection and evaluation strategy. Table~\ref{tab:service_domains} summarizes this mapping across representative domains.

\begin{table}[t]
\centering
\caption{FAA capabilities in representative service-oriented domains.}
\label{tab:service_domains}
\small
\renewcommand{\arraystretch}{1.12}
\begin{tabular}{L{1.35cm} L{1.45cm} L{2.0cm} L{2.7cm}}
\toprule
\textbf{Domain} & \textbf{FAA role} & \textbf{Deployment mode} & \textbf{Primary operational demand} \\
\midrule
Education & Engagement and participation cues & Cloud and edge classroom services & Robust low-latency feedback under varied capture conditions \\
Healthcare & Screening and monitoring support & Mobile-clinical hybrid workflows & Privacy-preserving inference and dependable reliability \\
Smart environments & Context-aware safety signaling & Edge gateways and local automation & Fast alerting with low false alarms \\
Transport & Driver-state support & In-vehicle real-time pipelines & Stability under illumination and motion shifts \\
Assistive systems & Social-emotional interaction support & Mobile and mixed-reality services & Personalized adaptation with transparent behavior \\
\bottomrule
\end{tabular}
\end{table}

\section{Task-Specific Advances Through a Service Lens}
\label{sec:task_advances}
\begin{figure}[t]
\centering
\includegraphics[width=\columnwidth]{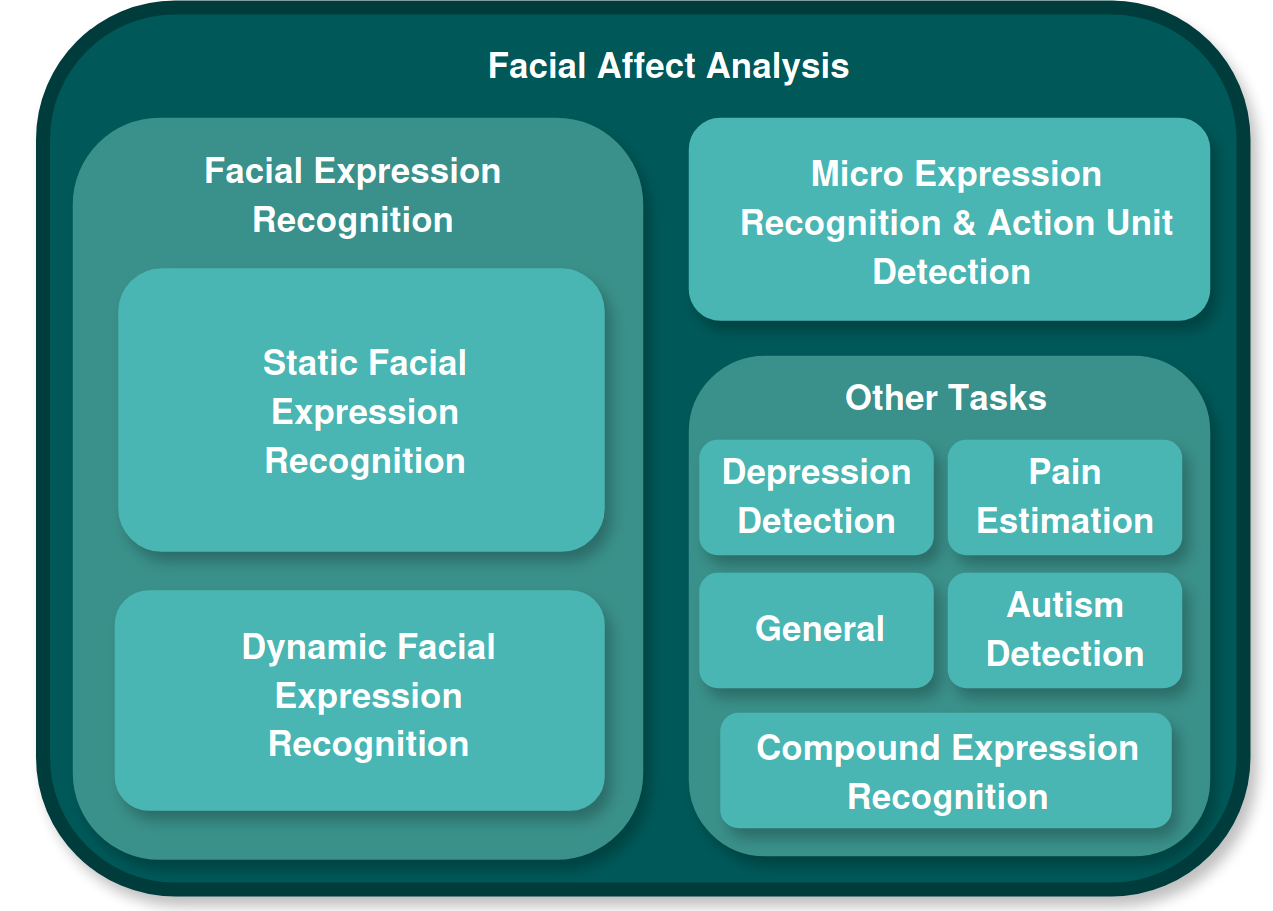}
\caption{FAA task families and their role in service-oriented analysis.}
\label{fig:section4_tasks}
\end{figure}

Figure~\ref{fig:section4_tasks} provides a compact visual map of FAA task families used throughout this section. We use it to connect method-level progress to service-level requirements, from low-latency static inference to temporally aware and fine-grained affect understanding.

\subsection{Static FER for low-latency services}
In Fig.~\ref{fig:section4_tasks}, static FER corresponds to the most deployment-ready task branch for real-time adaptation. Static FER remains the most practical entry point for service integration because the inference pipeline is compact and easier to optimize for edge hardware. Methods such as Ada-CM and DSAN reflect broader trends toward better supervision and discriminative feature learning under noisy labels \cite{li2022towards,fan2020facial}. For service design, this means static FER can support responsive adaptation loops in online education, kiosk interfaces, and customer-facing systems where frame-level reactivity is more important than long temporal context.

However, static FER has limitations that become visible at deployment time. First, single-frame predictions may be unstable under transient head motion, expression ambiguity, or background changes. Second, bias and class imbalance can produce uneven service behavior across users. Third, a model optimized for public benchmarks may overfit annotation conventions that differ from real service conditions. These factors motivate confidence-aware output handling and temporal smoothing at the service layer.

\subsection{Dynamic FER for continuity and trend detection}
The dynamic FER branch in Fig.~\ref{fig:section4_tasks} highlights the transition from instantaneous inference to sequence-aware service behavior. Dynamic FER models better align with services that depend on affective trends rather than instantaneous snapshots. Sequence-aware methods have improved temporal consistency and multi-scale motion understanding \cite{liu2023expression,xu2024multiscale}. In educational services, this supports sustained engagement estimation rather than isolated reactions. In transportation workflows, it supports monitoring readiness and distraction patterns across continuous driving intervals.

The trade-off is operational cost. Dynamic pipelines require buffering, synchronization, and higher memory footprints. They also demand stable frame-rate and capture quality, which can be difficult in mobile and edge contexts. Service engineers therefore face a strategic decision: adopt dynamic FER where trend fidelity matters, but reserve static FER or hybrid fallback paths where resource predictability is critical.

\subsection{MER and AU analysis for fine-grained services}
The fine-grained task branch in Fig.~\ref{fig:section4_tasks} captures MER and AU analysis, which target subtle cues that are often invisible to coarse classifiers. Micro-expression and AU-focused approaches are attractive in applications where subtle affective cues carry high semantic value, including wellbeing support and assistive interaction \cite{li2022deep,kollias2023multi}. Their potential strength is richer interpretability, because action-unit evidence can support more transparent service decisions than direct emotion labels.
Yet this potential comes with practical constraints. MER datasets remain relatively small and constrained, model sensitivity to capture quality is high, and low-amplitude facial motion is easily corrupted by environmental noise. As a result, these methods are best treated as high-value, domain-targeted components rather than universal service defaults. A robust system design can combine coarse static or dynamic FER for broad monitoring with optional AU-centered analysis in high-confidence windows.

\subsection{Architecture implications for service engineering}
From a deployment perspective, architecture families should be selected by operational fit. CNN pipelines are often sufficient for real-time edge use. Transformer and hybrid models can improve robustness to appearance variation, but may require cloud support or hardware acceleration \cite{xue2022vision,zheng2023poster}. Graph-based approaches can add structure-aware reasoning in AU-heavy settings but may increase system complexity \cite{liu2022graph}. The key insight is that architecture decisions should follow service-level constraints first, then benchmark ranking.

\section{FAA in Service-Oriented Applications}
\label{sec:applications}
\subsection{Education services}
Affective learning systems use FAA to estimate engagement and adapt teaching support. Lightweight FER pipelines have shown practical viability in online settings where responsiveness and continuity matter \cite{savchenko2022classifying}. Cloud-connected classroom systems extend this model by aggregating group-level signals for instructional analytics \cite{guo2022facial}.

From a deployment perspective, education is often the first domain where FAA is scaled because intervention loops are clear: detect engagement patterns, estimate attention drift, and adjust pacing or prompts. Three issues dominate real operation. First, engagement should not be mapped one-to-one to a single expression label, since students can appear neutral while remaining cognitively engaged. Second, classroom capture quality is highly variable, so confidence-aware gating is needed to avoid unstable adaptation. Third, fairness auditing and human override paths are essential for high-impact educational decisions.

\subsection{Healthcare and wellbeing services}
Healthcare-oriented FAA increasingly appears in mobile and blended monitoring workflows. Systems such as FacePsy show how on-device affect and behavior analysis can support wellbeing assessment while limiting sensitive data exposure \cite{islam2024facepsy}. Assistive platforms for neurodiverse users use FAA to scaffold social-emotional learning in caregiver-centered loops \cite{lyu2024emooly,lyu2024dailyconnect}. These systems demonstrate that service utility improves when FAA is embedded in human-in-the-loop workflows rather than treated as autonomous diagnosis.

Deployment in this domain must balance utility with sensitivity. Edge-first inference improves privacy, but constrains model complexity and update cadence. Hybrid designs are therefore common: immediate inference is local, while aggregated model refinement remains auditable. In practice, healthcare services should report intervention stability and false escalation behavior, since predictable support actions can be more important than marginal benchmark gains.

\subsection{Smart-home and safety services}
In smart environments, FAA contributes contextual signals for alerting and adaptive assistance. Home safety frameworks illustrate this pattern, but also expose governance risks such as false positives, unfair behavior across user groups, and consent boundaries \cite{kaushik2022isecurehome}. Service design should therefore include threshold governance, user override options, and audit trails.

Deployment deep dives in this domain show a recurring trade-off between responsiveness and conservative policy behavior. False positives can quickly erode trust and produce alert fatigue, while household heterogeneity makes one-size-fits-all models brittle. Robust systems require confirmation logic, personalization profiles, and uncertainty-aware conflict handling in multi-user scenes. Public-facing scenarios additionally require strict purpose limitation and data minimization to remain socially and legally acceptable.

\subsection{Transportation services}
In intelligent transport systems, FAA contributes to driver readiness and takeover support when combined with contextual signals. Recent multimodal approaches show promise for safety-critical adaptation, but only if robustness is preserved under changing real driving conditions \cite{gu2024emotake,saadi2024driver}. The most effective designs treat FAA as one signal among many and avoid single-channel dependence.

Deployment deep dives indicate that lighting variation, camera vibration, and head-pose change demand defensive temporal modeling and staged confidence policies. Low-confidence outputs should trigger confirmation windows instead of immediate intervention. Explainability is also critical under safety and liability constraints, so systems should log confidence and quality metadata alongside downstream policy actions. Continuous recalibration is required because route, season, and user behavior drift over time.

\subsection{Cross-domain lessons}
Despite domain differences, several lessons are consistent. First, FAA is most effective when treated as one component in a broader service loop, not as a stand-alone predictor. Second, uncertainty handling is as important as base accuracy. Third, deployment governance must be designed early, not retrofitted after technical integration. Fourth, service-aware evaluation should prioritize stability of interventions over single-frame correctness.

These lessons reinforce the central argument of this paper: progress in FAA methods becomes operationally meaningful only when translated into robust, transparent, and policy-aware service design.

\subsection{Service-oriented deployment checklist}
To translate these lessons into repeatable engineering practice, teams can use a compact service-oriented checklist before production rollout. The goal is not to maximize standalone recognition metrics, but to ensure that FAA behavior remains reliable when embedded in end-to-end service policies.

\begin{itemize}
    \item \textbf{Signal role definition:} Specify exactly how FAA influences decisions (advisory, gating, or triggering), and define safe fallback behavior when confidence is low.
    \item \textbf{Confidence and quality policy:} Bind prediction confidence and input-quality indicators to intervention aggressiveness, not only to logging.
    \item \textbf{Human oversight design:} Keep override and escalation paths explicit for high-impact actions in education, healthcare, safety, and transport.
    \item \textbf{Governance and compliance mapping:} Align data minimization, retention, and consent boundaries with domain regulations before deployment.
    \item \textbf{Drift and recalibration operations:} Plan periodic recalibration, subgroup monitoring, and post-deployment auditing as first-class service operations.
\end{itemize}

This checklist reframes FAA deployment as a lifecycle responsibility shared by model engineers, service architects, and domain operators.

\section{Deployment Architectures, Challenges, and Evaluation Protocols in Service-Oriented Software Ecosystems (SoSE)}
\label{sec:deployment_patterns}
\subsection{Cloud-centric orchestration}
Cloud-centric pipelines support model complexity, centralized updates, and large-scale analytics. This pattern is appropriate for institutional services such as classrooms and managed care platforms where network connectivity and governance infrastructure are strong \cite{guo2022facial}. However, cloud-centric inference increases dependence on communication quality and raises data handling concerns that can reduce user trust.

\subsection{Edge-first and on-device operation}
Edge-first designs move inference closer to the user. This reduces round-trip latency and can improve privacy by limiting raw video transfer. Mobile wellbeing and assistive scenarios especially benefit from this pattern \cite{islam2024facepsy,lyu2024dailyconnect}. The main challenge is maintaining model quality under restricted compute and thermal budgets.

\subsection{Hybrid federation between edge and cloud}
Many realistic systems require hybrid execution: fast local inference for immediate interaction and cloud services for heavier model updates, long-horizon analytics, or policy optimization. Hybrid designs can improve resilience, but require explicit consistency management between local and centralized components. In addition, fallback behavior must be well defined for connectivity drops or uncertain model outputs.

\subsection{Interoperable FAA services}
To become reusable service modules, FAA components should expose structured outputs beyond class labels. Useful interfaces include confidence, uncertainty bands, latency metadata, and quality flags. These outputs allow downstream services to apply adaptive policies, such as conservative responses under uncertainty, escalation to human oversight, or fusion with complementary sensing channels.

A practical SoSE interpretation is to package FAA as a versioned service component with explicit interface contracts. The input contract should document modality assumptions (single frame, short sequence, or multimodal context), capture constraints, and acceptable quality thresholds. The output contract should include not only affect predictions, but calibrated confidence, abstention states, and machine-readable error conditions for policy engines. Operational contracts should also declare latency envelopes (for example p95 inference delay), expected availability, and fallback modes under degraded sensing or network loss. This contract-first design improves interoperability across independently evolving services and reduces brittle integrations when FAA models are updated.

\subsection{Main deployment challenges in SoSE setups}
When FAA is deployed as part of service-oriented software ecosystems, three challenge groups dominate. First, \textbf{data quality and domain shift}: label noise, class imbalance, sensor differences, and changing usage patterns can destabilize adaptation behavior even when benchmark results are strong \cite{li2022towards,zeng2022face2exp}. Second, \textbf{robustness and runtime constraints}: occlusion, lighting variation, and motion artifacts interact with latency and energy limits, especially in edge and mobile deployments \cite{xue2022vision,liu2023expression,xu2024multiscale,chen2024static}. Third, \textbf{governance and trust}: fairness gaps and unclear privacy boundaries can reduce user acceptance and create operational risk \cite{dominguez2024metrics,islam2024facepsy}.

These challenges imply that architecture choice in SoSE is not a one-time model selection step. It is a lifecycle decision that must couple model updates, fallback policy, and auditability across edge-cloud execution.

\subsection{Evaluation protocols for FAA in SoSE}
Proper evaluation in SoSE should go beyond standalone recognition scores and test whether FAA improves service behavior reliably. Three protocol layers are essential.

\textbf{Protocol layer 1: dataset and context realism.}
Evaluation sets should match deployment conditions, including camera quality, compression artifacts, and user-behavior variability. Cross-domain validation should be treated as a default requirement, not an optional ablation, to expose hidden domain assumptions.

\textbf{Protocol layer 2: temporal and intervention stability.}
Because services consume FAA outputs over time, protocols should measure prediction drift, confidence fluctuation, and policy-trigger consistency. This reveals whether a model supports stable adaptation or causes contradictory service actions.

\textbf{Protocol layer 3: governance-aware operational reporting.}
Each study should report where inference runs, what data are retained, how uncertainty is handled, and how fairness is monitored during operation. This makes results transferable across SoSE deployments and supports accountable integration.

\subsection{SoSE evaluation workflow and reporting template}
To make protocol execution reproducible, SoSE studies should follow a compact workflow that links model validation to service behavior.

\textbf{Step 1: define the service objective and decision scope.}
Specify whether FAA output is advisory, gating, or triggering, and identify which downstream service actions are affected.

\textbf{Step 2: profile deployment context and runtime envelope.}
Document camera setup, compute target, network assumptions, and latency budget for the intended edge, cloud, or hybrid architecture.

\textbf{Step 3: execute multi-context validation.}
Evaluate in-distribution and cross-domain settings, then report degradation patterns under realistic perturbations such as lighting shifts, compression, pose variation, and partial occlusion.

\textbf{Step 4: evaluate temporal intervention behavior.}
Measure not only prediction quality but also service policy stability over time, including false escalation tendency, contradictory adaptation events, and abstention behavior under low confidence.

\textbf{Step 5: audit governance outcomes.}
Report subgroup fairness behavior, data retention pathways, and escalation mechanisms for high-impact use cases.

\textbf{Step 6: publish an operational report card.}
Summarize predictive quality, runtime metrics, fairness indicators, privacy mode, and intervention stability in a single deployment-facing report. This report card should accompany benchmark tables so that integration teams can directly assess SoSE readiness.

\subsection{Lifecycle operations for FAA services in SoSE}
Evaluation should continue after initial deployment as part of a managed service lifecycle. A practical loop is: deploy with explicit policy thresholds, monitor confidence and intervention behavior in production, trigger recalibration when drift signals exceed predefined bounds, and maintain rollback-ready model versions for safe recovery. In this loop, observability is central: teams should track confidence decay, contradiction rate between consecutive interventions, subgroup reliability trends, and latency-percentile regressions under realistic load. Governance checks should be integrated with the same cadence, so privacy mode, retention assumptions, and escalation policies are audited alongside model quality. Treating FAA as a continuously operated service, rather than a one-time model artifact, is essential for dependable SoSE integration.

\subsection{Service-aware evaluation dimensions}
Benchmark accuracy is necessary but insufficient for FAA-enabled services. Practical evaluation should also assess robustness to distribution shift, end-to-end latency, fairness across user groups, privacy of data flows, and stability under noisy input. For SoSE reporting, two useful KPIs are p95 sensing-to-actuation latency and contradictory-intervention rate. Together, these dimensions shift evaluation from model ranking to service reliability, as operationalized in Table~\ref{tab:evaluation_dimensions}.

\begin{table*}[t]
\centering
\caption{Recommended evaluation dimensions for FAA-enabled services.}
\label{tab:evaluation_dimensions}
\small
\renewcommand{\arraystretch}{1.15}
\begin{tabular}{L{2.3cm} L{3.1cm} L{4.0cm} L{4.0cm}}
\toprule
\textbf{Dimension} & \textbf{What to measure} & \textbf{Why it matters in services} & \textbf{Typical mitigation strategy} \\
\midrule
Predictive quality & Class performance and calibration & Core affect inference validity & Balanced training and uncertainty calibration \\
Robustness under shift & Performance across lighting, pose, device, and context changes & Prevent unstable adaptation behavior & Domain-robust training and stress testing \\
Runtime behavior & End-to-end latency and throughput under target hardware & Determines real-time usability & Model compression, staged inference, edge acceleration \\
Fairness behavior & Group-level performance consistency & Avoid unequal service impact & Bias auditing and re-weighted optimization \\
Privacy posture & Data residency and exposure pathways & Governs user trust and compliance & On-device inference and minimal data retention \\
Intervention stability & Consistency of service actions under noisy affect input & Prevent erratic service responses & Temporal smoothing and policy safeguards \\
\bottomrule
\end{tabular}
\end{table*}

\section{Implementation Guidelines for FAA-Enabled Services}
\label{sec:implementation_guidelines}
The hybrid service perspective suggests a practical implementation playbook. The following guidelines summarize repeatable design choices that improve reliability without requiring a single universal model architecture.

\begin{enumerate}
    \item \textbf{Separate sensing from intervention logic:} FAA modules should provide probabilistic affect signals, while intervention policies remain explicit and domain-controlled. This separation improves auditability and allows policy updates without retraining the entire affect model. In sensitive domains, it also makes human oversight easier because decisions can be traced to both model output and policy constraints.
    \item \textbf{Use confidence-aware adaptation:} Service behavior should scale with confidence. High-confidence predictions can trigger direct adaptation, while low-confidence predictions should route to conservative actions, delayed decisions, or multimodal confirmation. This strategy reduces unstable service behavior under noisy capture conditions and aligns with trust-aware operation.
    \item \textbf{Design for degraded operation:} Production services must remain useful under partial failure. Examples include reduced frame rate, temporary occlusion, missing sensors, or network interruptions. Degraded modes can rely on lightweight fallback models, slower adaptation cadence, or user-confirmed actions. This design principle is particularly important for mobile and vehicle services where conditions change rapidly.
    \item \textbf{Align architecture with hardware envelope:} Model selection should begin with target hardware limits. Edge-first deployments often favor compact CNN or distilled hybrids; cloud-assisted deployments can support heavier temporal models when latency budgets allow \cite{chen2024static,xu2024multiscale}. This hardware-first planning prevents overfitting the service stack to models that cannot sustain real-time operation.
    \item \textbf{Monitor drift continuously:} Service populations and environments evolve. A model that works at launch may drift as camera placement, user behavior, or interaction design changes. Continuous monitoring should track distribution shift indicators, confidence decay, and intervention error signatures. When drift is detected, retraining or calibration updates should be applied with fairness and privacy constraints preserved.
    \item \textbf{Integrate multimodal evidence selectively:} Multimodal fusion can improve robustness, but unnecessary fusion can increase complexity and failure surfaces. A pragmatic approach is selective fusion: keep FAA as the primary channel where visual quality is strong, and activate additional signals only when uncertainty rises. Transportation and wellbeing services already demonstrate the value of this staged strategy \cite{gu2024emotake,islam2024facepsy}.
    \item \textbf{Document assumptions for reproducible deployment:} Service-oriented FAA studies should report operational assumptions in detail: camera setup, compute profile, policy thresholds, privacy mode, and fallback behavior. Without this documentation, reproduction remains benchmark-level only and deployment lessons cannot transfer effectively across domains.
\end{enumerate}

\subsection{Domain playbooks for implementation}
Beyond generic guidance, deployment teams benefit from domain-specific playbooks that map FAA outputs to concrete service actions. Four recurring playbook patterns appear across recent systems.

\textbf{Education playbook.}
Use short temporal windows to estimate engagement trends, not isolated frame labels. Couple these trends with low-risk adaptation actions first (pace adjustment, optional prompts, content modality switching), and reserve high-impact actions for sustained evidence. In classroom dashboards, expose uncertainty and quality indicators so educators can interpret model output in context. This pattern preserves pedagogical flexibility while reducing overreaction to transient affect signals.

\textbf{Healthcare and wellbeing playbook.}
Prioritize edge inference for immediate support and privacy, sending only compact, policy-approved summaries to the cloud for longitudinal tracking. FAA outputs should be treated as risk indicators, not diagnoses, with clear escalation to caregivers or clinicians when confidence is low or concern persists. This improves acceptance while aligning technical behavior with clinical accountability.

\textbf{Smart environments playbook.}
Treat FAA as contextual enrichment for safety or comfort automation, not as an autonomous trigger. Combine affect cues with scene state and interaction history, then require confirmation logic before high-salience alerts. Provide household-level controls for sensitivity and override settings, and keep audit traces for major automated actions. This approach reduces alert fatigue while maintaining user trust.

\textbf{Transportation playbook.}
Fuse FAA with gaze, posture, and telemetry in staged pipelines. Low-confidence events should prompt monitoring, while persistent multi-signal evidence can trigger stronger interventions. Log confidence, input quality, and rationale, since traceability is critical in safety-sensitive settings.

These playbooks emphasize a shared principle: service value emerges when FAA is connected to explicit policy design, bounded intervention authority, and continuous operational monitoring.

\subsection{Worked example: smart city and smart-home deployment}
Consider a smart-city safety service that integrates neighborhood smart-home hubs, public kiosks, and municipal coordination dashboards. FAA modules at the edge estimate affective distress and interaction state, while a separate policy layer decides whether to trigger low-risk local interventions (for example, adaptive lighting or voice assistance) or escalate to human operators (Guidelines 1--2). When cameras are partially occluded, nighttime illumination is poor, or network links degrade, the system switches to conservative degraded behavior with slower update cadence and confirmation checks instead of aggressive alerts (Guideline 3).

\begin{figure}[t]
\centering
\includegraphics[width=\columnwidth]{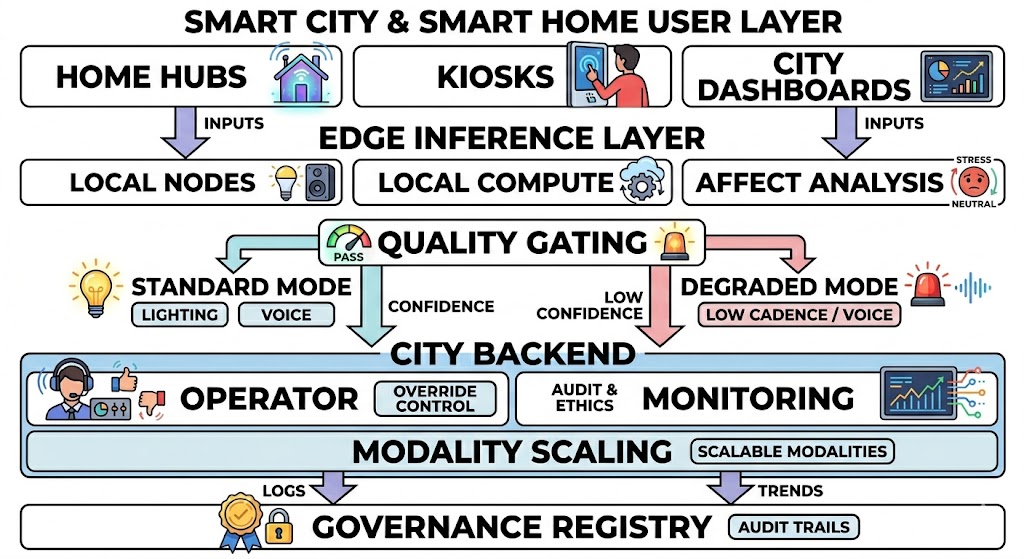}
\caption{Smart city and smart-home deployment example for applying Section~\ref{sec:implementation_guidelines} guidelines.}
\label{fig:section7_example}
\end{figure}

Figure~\ref{fig:section7_example} illustrates this deployment flow, including edge inference, confidence-gated policy decisions, degraded fallback behavior, and city-level monitoring and audit loops.

Because residential gateways and public edge nodes have limited compute, compact models handle real-time inference locally, while city backends aggregate anonymized trends for planning and service tuning (Guideline 4). Operational monitoring tracks confidence drift, subgroup impact, and false-alert patterns across districts to surface reliability and fairness issues early (Guideline 5). Additional modalities such as IoT context signals or interaction logs are activated only under elevated uncertainty, limiting unnecessary complexity in stable conditions (Guideline 6). Finally, deployment documentation records device profiles, threshold policies, retention limits, fallback logic, and escalation responsibilities so the service remains auditable and reproducible across smart-home and smart-city environments (Guideline 7). Table~\ref{tab:deployment_guidelines} condenses these implementation rules into a deployment checklist.

\begin{table*}[t]
\centering
\caption{Practical guidelines for deploying FAA in service-oriented systems.}
\label{tab:deployment_guidelines}
\small
\renewcommand{\arraystretch}{1.25}
\begin{tabular}{L{2.3cm} L{4.1cm} L{3.8cm} L{3.6cm}}
\toprule
\textbf{Guideline} & \textbf{Implementation idea} & \textbf{Primary benefit} & \textbf{Typical risk if ignored} \\
\midrule
Sensing-policy separation & Keep FAA inference and intervention policies as distinct modules & Better auditability and safer updates & Opaque decisions and brittle service behavior \\
Confidence-aware adaptation & Gate interventions by confidence and quality signals & Reduces unstable actions under noise & Frequent contradictory interventions \\
Degraded operation modes & Define fallback behavior for capture or network failures & Improves service continuity & Hard failures in real contexts \\
Hardware-aligned architecture & Select model families by latency and memory envelope & Sustainable real-time operation & Models too heavy for deployment \\
Drift monitoring & Track distribution and confidence changes over time & Early detection of reliability decay & Silent performance collapse \\
Selective multimodal fusion & Add channels only when uncertainty justifies complexity & Better robustness with manageable cost & Over-complex systems with fragile integration \\
Deployment documentation & Report compute, policy, privacy, and fallback assumptions & Transferable and reproducible evidence & Non-reproducible deployment claims \\
\bottomrule
\end{tabular}
\end{table*}

\section{Future Visions and Open Problems}
\label{sec:roadmap}
FAA’s service-oriented transition creates a research agenda beyond benchmarks, requiring joint progress in model quality, service behavior, and governance readiness. Key open problems are summarized below.

\begin{itemize}
    \item \textbf{Robust generalization under operational shift:} Most FAA models still degrade when device characteristics, camera placement, or user behavior change. Domain adaptation methods have improved transfer behavior, but practical services need continuous robustness under non-stationary conditions \cite{li2022towards,zeng2022face2exp}. A promising direction is lifecycle-aware learning, where models are evaluated and updated as part of service operation rather than as one-time static artifacts.
    \item \textbf{Trustworthy uncertainty handling:} FAA-enabled services need reliable uncertainty signals that are actionable in policy logic. Current systems often expose softmax confidence without calibration guarantees, which can lead to unstable interventions. Future work should focus on calibrated uncertainty, quality-aware gating, and explicit abstention behavior when evidence is unreliable.
    \item \textbf{Fairness-preserving adaptation:} Fairness is difficult in long-running services because user distributions evolve over time. A model that appears balanced at deployment can drift toward uneven outcomes. Research should explore fairness monitoring integrated with drift detection and adaptation workflows, so that updates improve robustness without amplifying subgroup disparities \cite{dominguez2024metrics}.
    \item \textbf{Efficient multimodal orchestration:} Multimodal fusion improves robustness, but naive fusion increases system complexity and latency. Open challenges include dynamic modality selection, adaptive fusion based on confidence, and graceful degradation when one channel fails. Transportation and wellbeing scenarios provide strong motivation for this direction \cite{gu2024emotake,islam2024facepsy}.
    \item \textbf{Reproducible deployment evidence:} Many papers still report model performance without sufficient deployment context. Service-oriented FAA research should standardize reporting of hardware profiles, runtime behavior, privacy mode, fallback policies, and intervention metrics. This would make findings transferable across domains and reduce the gap between experimental claims and production outcomes.
    \item \textbf{Human oversight in high-impact services:} In healthcare, education, and safe transport scenarios, FAA should support human decision makers rather than silently replacing them. Open problems include interface design for uncertainty communication, escalation policy design, and accountability workflows for model-assisted decisions.
\end{itemize}

Table~\ref{tab:roadmap} maps these open problems to near-term priorities and expected service impact.

\begin{table}[t]
\centering
\caption{Roadmap priorities for service-oriented FAA research.}
\label{tab:roadmap}
\small
\renewcommand{\arraystretch}{1.12}
\begin{tabular}{L{1.9cm} L{2.2cm} L{2.9cm}}
\toprule
\textbf{Priority} & \textbf{Near-term goal} & \textbf{Service impact} \\
\midrule
Robust transfer & Improve cross-context consistency & Fewer adaptation failures in real operation \\
Uncertainty calibration & Enable reliable abstention and gating & Safer intervention policies \\
Fairness lifecycle & Monitor equity during updates & More trustworthy user experience \\
Efficient multimodality & Select channels adaptively & Better robustness with bounded cost \\
Deployment reporting & Standardize operational metrics & Reproducible and transferable evidence \\
Human oversight & Design accountable interaction loops & Better acceptance in sensitive domains \\
\bottomrule
\end{tabular}
\end{table}

\subsection{Service-level outlook for FAA-enabled services}
\label{sec:future_directions}
The next stage of FAA research should prioritize service-level dependability over isolated benchmark gains. Four transitions appear most likely: privacy-preserving edge-first pipelines in sensitive domains, broader multimodal fusion with contextual signals \cite{gu2024emotake,islam2024facepsy}, interoperable FAA interfaces that expose uncertainty and failure states, and operational reporting that captures behavior under real deployment shift.

Continuous adaptation with safeguards is also essential. Because service pipelines operate over long horizons and changing populations, models should support monitored updates while preserving fairness constraints, auditability, and explicit human oversight for high-impact decisions.

From an architectural perspective, future systems will likely move toward composable FAA services rather than monolithic models. In this view, expression estimation, confidence calibration, context fusion, and intervention policy support are implemented as separate but interoperable components. Such modularization enables targeted upgrades, safer rollback strategies, and clearer accountability pathways when failures occur. It also allows different domains to share core FAA components while customizing policy logic for local constraints.
Another likely direction is stronger alignment between model training and service objectives. Instead of optimizing only recognition loss, future approaches can include service-level signals such as intervention stability, low-confidence abstention quality, and fairness consistency over time. This broadens optimization from static prediction quality to operational utility. In practice, this could reduce the gap between laboratory model ranking and real service performance.

Future FAA-enabled services should use clear transparency contracts explaining what is sensed, how long data is kept, and how users can override affect-aware behavior. This can support compliance while improving trust, adoption, and user control. These directions position FAA as dependable middleware for adaptive service ecosystems.

\section{Conclusions}
\label{sec:conclusions}
This paper revised the FAA survey perspective by following a hybrid strategy: preserving FAA methodological depth while reframing the field around service-oriented deployment. We showed that recent FAA advances are most meaningful when interpreted through operational constraints, including robustness, latency, fairness, privacy, and integration effort.
FAA can already support adaptive services in education, healthcare, smart environments, transport, and assistive computing. However, dependable adoption depends on service-aware evaluation and architectures suited to each deployment context.

This conclusion has clear scope limits. The review prioritizes service-oriented synthesis over exhaustive method cataloging, focuses on representative work from the last decade, and relies on studies with varying deployment detail. Transferability also remains limited across domains with different risks and regulations. As datasets, benchmarks, and reporting practices evolve, recommendations will need refinement, particularly for fairness monitoring, uncertainty calibration, and policy-aware optimization. Despite these limits, the review offers practical value by identifying FAA capabilities ready for integration, common deployment constraints, and research directions that can improve operational robustness.

In synthesis, FAA is moving from model-centric benchmarking toward service-critical capability engineering. Progress will depend on coupling methodological innovation with deployment realism: edge-cloud architectural fit, uncertainty-aware orchestration, fairness and privacy governance, and evaluation protocols centered on intervention stability. This integrated perspective is a practical pathway toward dependable FAA-enabled SoSE deployments.
For Service-Oriented Systems Engineering communities, the main contribution is this integration view: FAA should be designed, evaluated, and governed as an operational service component with explicit contracts, measurable runtime guarantees, and lifecycle accountability across evolving service ecosystems.

\section*{Acknowledgments}
This work has received funding from the European Union's Horizon Europe research and innovation programme under Grant Agreement No. 101168042 project TRIFFID (auTonomous Robotic aId For increasing First responders Efficiency) and No. 101189557 project TORNADO (foundaTion mOdels for Robots that haNdle smAll, soft and Deformable Objects).

\bibliographystyle{ieeetr}
\bibliography{references}

@article{page2021prisma,
  title={The PRISMA 2020 statement: an updated guideline for reporting systematic reviews},
  author={Page, Matthew J and McKenzie, Joanne E and Bossuyt, Patrick M and Boutron, Isabelle and Hoffmann, Tammy C and Mulrow, Cynthia D and Shamseer, Larissa and Tetzlaff, Jennifer M and Akl, Elie A and Brennan, Sue E and others},
  journal={bmj},
  volume={372},
  year={2021},
  publisher={British Medical Journal Publishing Group}
}

@article{savchenko2022classifying,
  title={Classifying emotions and engagement in online learning based on a single facial expression recognition neural network},
  author={Savchenko, Andrey V and Savchenko, Lyudmila V and Makarov, Ilya},
  journal={IEEE Transactions on Affective Computing},
  volume={13},
  number={4},
  pages={2132--2143},
  year={2022},
  publisher={IEEE}
}

@article{kaushik2022isecurehome,
  title={iSecureHome: A deep fusion framework for surveillance of smart homes using real-time emotion recognition},
  author={Kaushik, Harshit and Kumar, Tarun and Bhalla, Kriti},
  journal={Applied Soft Computing},
  volume={122},
  pages={108788},
  year={2022},
  publisher={Elsevier}
}

@article{liu2022graph,
  title={Graph-based facial affect analysis: A review},
  author={Liu, Yang and Zhang, Xingming and Li, Yante and Zhou, Jinzhao and Li, Xin and Zhao, Guoying},
  journal={IEEE Transactions on Affective Computing},
  volume={14},
  number={4},
  pages={2657--2677},
  year={2022},
  publisher={IEEE}
}

@article{li2020deep,
  title={Deep facial expression recognition: A survey},
  author={Li, Shan and Deng, Weihong},
  journal={IEEE transactions on affective computing},
  volume={13},
  number={3},
  pages={1195--1215},
  year={2020},
  publisher={IEEE}
}

@article{li2022deep,
  title={Deep learning for micro-expression recognition: A survey},
  author={Li, Yante and Wei, Jinsheng and Liu, Yang and Kauttonen, Janne and Zhao, Guoying},
  journal={IEEE Transactions on Affective Computing},
  volume={13},
  number={4},
  pages={2028--2046},
  year={2022},
  publisher={IEEE}
}

@inproceedings{li2022towards,
  title={Towards semi-supervised deep facial expression recognition with an adaptive confidence margin},
  author={Li, Hangyu and Wang, Nannan and Yang, Xi and Wang, Xiaoyu and Gao, Xinbo},
  booktitle={Proceedings of the IEEE/CVF conference on computer vision and pattern recognition},
  pages={4166--4175},
  year={2022}
}

@article{fan2020facial,
  title={Facial expression recognition with deeply-supervised attention network},
  author={Fan, Yingruo and Li, Victor OK and Lam, Jacqueline CK},
  journal={IEEE transactions on affective computing},
  volume={13},
  number={2},
  pages={1057--1071},
  year={2020},
  publisher={IEEE}
}

@article{liu2023expression,
  title={Expression snippet transformer for robust video-based facial expression recognition},
  author={Liu, Yuanyuan and Wang, Wenbin and Feng, Chuanxu and Zhang, Haoyu and Chen, Zhe and Zhan, Yibing},
  journal={Pattern Recognition},
  volume={138},
  pages={109368},
  year={2023},
  publisher={Elsevier}
}

@inproceedings{zeng2022face2exp,
  title={Face2exp: Combating data biases for facial expression recognition},
  author={Zeng, Dan and Lin, Zhiyuan and Yan, Xiao and Liu, Yuting and Wang, Fei and Tang, Bo},
  booktitle={Proceedings of the IEEE/CVF conference on computer vision and pattern recognition},
  pages={20291--20300},
  year={2022}
}

@article{xue2022vision,
  title={Vision transformer with attentive pooling for robust facial expression recognition},
  author={Xue, Fanglei and Wang, Qiangchang and Tan, Zichang and Ma, Zhongsong and Guo, Guodong},
  journal={IEEE Transactions on Affective Computing},
  volume={14},
  number={4},
  pages={3244--3256},
  year={2022},
  publisher={IEEE}
}

@inproceedings{zheng2023poster,
  title={Poster: A pyramid cross-fusion transformer network for facial expression recognition},
  author={Zheng, Ce and Mendieta, Matias and Chen, Chen},
  booktitle={Proceedings of the IEEE/CVF International Conference on Computer Vision},
  pages={3146--3155},
  year={2023}
}

@article{guo2022facial,
  title={Facial expressions recognition with multi-region divided attention networks for smart education cloud applications},
  author={Guo, Yifei and Huang, Jian and Xiong, Mingfu and Wang, Zhongyuan and Hu, Xinrong and Wang, Jihong and Hijji, Mohammad},
  journal={Neurocomputing},
  volume={493},
  pages={119--128},
  year={2022},
  publisher={Elsevier}
}

@inproceedings{kollias2023multi,
  title={Multi-label compound expression recognition: C-expr database \& network},
  author={Kollias, Dimitrios},
  booktitle={Proceedings of the IEEE/CVF conference on computer vision and pattern recognition},
  pages={5589--5598},
  year={2023}
}

@article{chen2024static,
  title={From static to dynamic: Adapting landmark-aware image models for facial expression recognition in videos},
  author={Chen, Yin and Li, Jia and Shan, Shiguang and Wang, Meng and Hong, Richang},
  journal={IEEE Transactions on Affective Computing},
  year={2024},
  publisher={IEEE}
}

@article{dominguez2024metrics,
  title={Metrics for dataset demographic bias: A case study on facial expression recognition},
  author={Dominguez-Catena, Iris and Paternain, Daniel and Galar, Mikel},
  journal={IEEE Transactions on Pattern Analysis and Machine Intelligence},
  volume={46},
  number={8},
  pages={5209--5226},
  year={2024},
  publisher={IEEE}
}

@article{xu2024multiscale,
  title={Multiscale facial expression recognition based on dynamic global and static local attention},
  author={Xu, Jie and Li, Yang and Yang, Guanci and He, Ling and Luo, Kexin},
  journal={IEEE Transactions on Affective Computing},
  year={2024},
  publisher={IEEE}
}

@article{islam2024facepsy,
  title={Facepsy: An open-source affective mobile sensing system-analyzing facial behavior and head gesture for depression detection in naturalistic settings},
  author={Islam, Rahul and Bae, Sang Won},
  journal={Proceedings of the ACM on Human-Computer Interaction},
  volume={8},
  number={MHCI},
  pages={1--32},
  year={2024},
  publisher={ACM New York, NY, USA}
}

@article{lyu2024emooly,
  title={EMooly: supporting autistic children in collaborative social-emotional learning with caregiver participation through interactive AI-infused and AR activities},
  author={Lyu, Yue and Liu, Di and An, Pengcheng and Tong, Xin and Zhang, Huan and Katsuragawa, Keiko and Zhao, Jian},
  journal={Proceedings of the ACM on Interactive, Mobile, Wearable and Ubiquitous Technologies},
  volume={8},
  number={4},
  pages={1--36},
  year={2024},
  publisher={ACM New York, NY, USA}
}

@article{gu2024emotake,
  title={EmoTake: Exploring Drivers’ Emotion for Takeover Behavior Prediction},
  author={Gu, Yu and Weng, Yibing and Wang, Yantong and Wang, Meng and Zhuang, Guohang and Huang, Jinyang and Peng, Xiaolan and Luo, Liang and Ren, Fuji},
  journal={IEEE Transactions on Affective Computing},
  volume={15},
  number={4},
  pages={2112--2127},
  year={2024},
  publisher={IEEE}
}

@article{lyu2024dailyconnect,
  title={DailyConnect: Piloting Interventions of Situation-Based Emotional Understanding in Naturalistic Home Settings for Children with Autism Spectrum Disorder},
  author={Lyu, Chengchen and Chen, Hui and Xu, Tong and Peng, Xiaolan and Huang, Faliang and Wang, Hongan},
  journal={International Journal of Human--Computer Interaction},
  volume={40},
  number={17},
  pages={4647--4660},
  year={2024},
  publisher={Taylor \& Francis}
}

@article{saadi2024driver,
  title={Driver’s facial expression recognition: A comprehensive survey},
  author={Saadi, Ibtissam and Taleb-Ahmed, Abdelmalik and Hadid, Abdenour and El Hillali, Yassin and others},
  journal={Expert Systems with Applications},
  volume={242},
  pages={122784},
  year={2024},
  publisher={Elsevier}
}

\end{document}